\pdfoutput=1

\documentclass[11pt]{article}

\usepackage[final]{acl}

\usepackage{times}
\usepackage{latexsym}

\usepackage[T1]{fontenc}

\usepackage[utf8]{inputenc}

\usepackage{microtype}

\usepackage{inconsolata}

\usepackage{graphicx}
\usepackage{tabularx}
\usepackage{booktabs}
\usepackage{adjustbox}
\usepackage{fontawesome}
\usepackage{amstext}
\usepackage{amsmath}
\usepackage{amssymb}
\usepackage{microtype}
\usepackage{multirow}
\usepackage{algorithm}
\usepackage{multicol}
\usepackage{algpseudocode}
\usepackage{amsfonts}
\usepackage{makecell}
\usepackage{colortbl} 
\usepackage{marvosym}
\usepackage{pifont}
\usepackage{hyperref}
\definecolor{mygreen}{RGB}{0,160,0}
\newcommand{\cmarkgreen}{\textcolor{mygreen}{{\Large \ding{51}}}}
\definecolor{myred}{RGB}{178,34,34}
\newcommand{\xmarkred}{\textcolor{myred}{{\Large \ding{55}}}}

\title{URL: Universal Referential Knowledge Linking via Task-instructed
Representation Compression}

\author{Zhuoqun Li${}^{1,4}$, Hongyu Lin${}^{1}$, Tianshu Wang${}^{1,4}$, Boxi Cao${}^{1,4}$, Yaojie Lu${}^{1}$, \\ \textbf{Weixiang Zhou${}^{1}$, \textbf{Hao Wang}${}^{3}$, \textbf{Zhenyu Zeng}${}^{3}$, \textbf{Le Sun}${}^{1,2}$, \textbf{Xianpei Han}${}^{1,2}$}\\
${}^{1}$Chinese Information Processing Laboratory ~ ${}^{2}$State Key Laboratory of Computer Science \\
Institute of Software, Chinese Academy of Sciences, Beijing, China\\
${}^{3}$Alibaba Cloud Intelligence Group\\
${}^{4}$University of Chinese Academy of Sciences, Beijing, China \\
{\tt \{lizhuoqun2021,hongyu,tianshu2020,boxi2020,luyaojie,weixiang\}@iscas.ac.cn} \\
cashenry@126.com~~zhenyu.zzy@alibaba-inc.com~~{\tt \{sunle,xianpei\}@iscas.ac.cn}}

\begin{document}
\maketitle

\begin{abstract}
Linking a claim to grounded references is a critical ability to fulfill human demands for authentic and reliable information. Current studies are limited to specific tasks like information retrieval or semantic matching, where the claim-reference relationships are unique and fixed, while the referential knowledge linking (RKL) in real-world can be much more diverse and complex. In this paper, we propose universal referential knowledge linking (URL), which aims to resolve diversified referential knowledge linking tasks by one unified model. To this end, we propose a LLM-driven task-instructed representation compression, as well as a multi-view learning approach, in order to effectively adapt the instruction following and semantic understanding abilities of LLMs to referential knowledge linking. Furthermore, we also construct a new benchmark to evaluate ability of models on referential knowledge linking tasks across different scenarios. Experiments demonstrate that universal RKL is challenging for existing approaches, while the proposed framework can effectively resolve the task across various scenarios, and therefore outperforms previous approaches by a large margin.

\end{abstract}

\section{Introduction}

Access to reliable, authentic, and well-founded information is a fundamental human necessity~\citep{clifford1877ethics, cole2011theory}. In recent years, with the rise of AI-generated content (AIGC), there has been an increasing demand for the verification of information authenticity and the identification of corresponding references~\citep{lewis2020retrieval, cao2023comprehensive, zhao2023survey}. Within this context, linking a claim to its grounded references, a task we refer to as \textit{\textbf{referential knowledge linking (RKL)}}, plays an indispensable role. Given a claim, the objective of referential knowledge linking is to associate it with relevant references within trustworthy information sources, thereby facilitating the verification of related information~\cite{yue2023disclawllm, huang2023survey, gao2023retrievalaugmented}. RKL is crucial for affirming the accuracy and validity of information, ensuring the reliability and legality of AIGC-produced content, and the efficient organization and management of information.

\begin{figure}[t!]
\centering
\includegraphics[width=\linewidth]{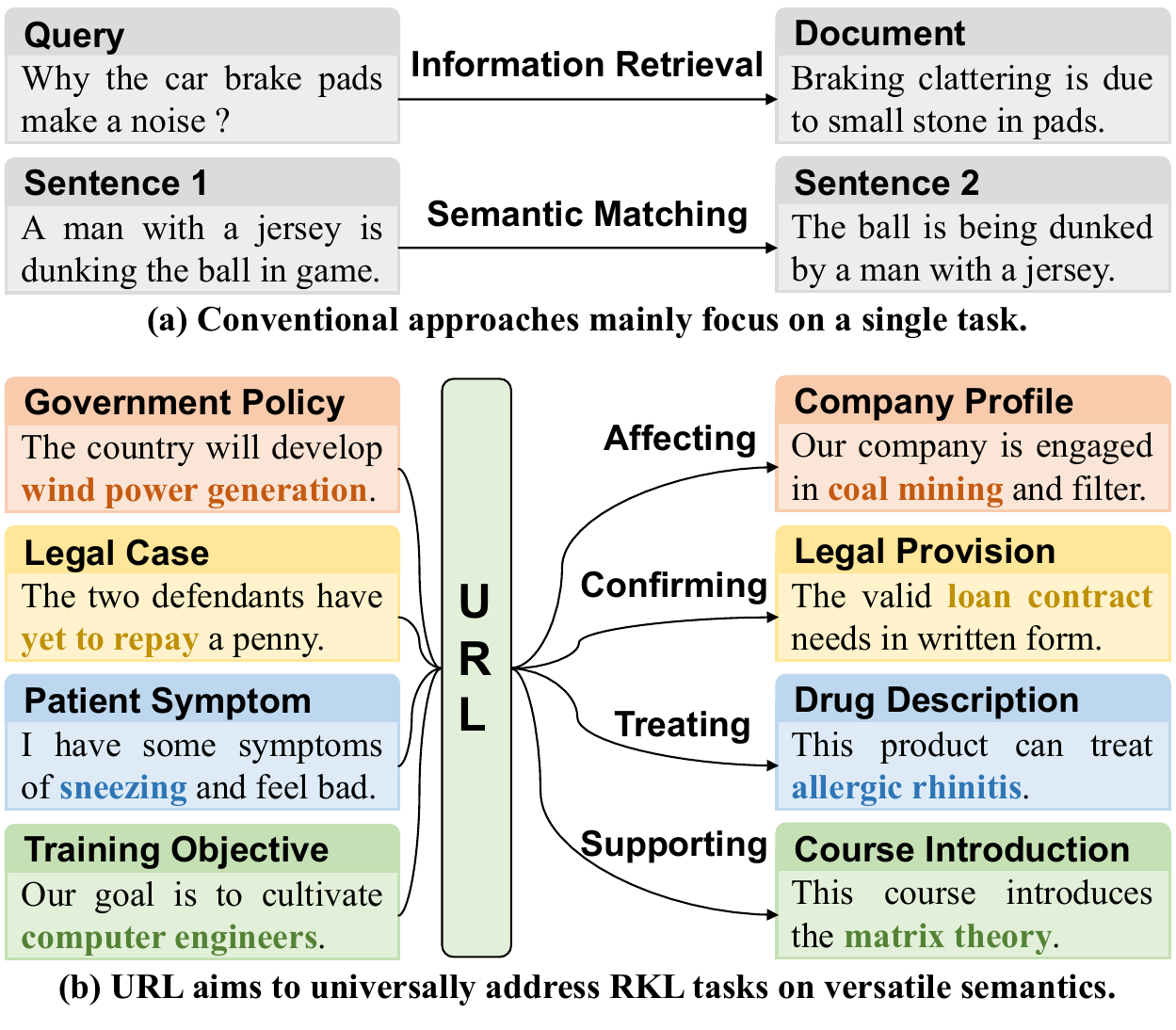} 
\caption{Compared to conventional approaches focus on a single task, URL aims to universally address RKL tasks on versatile semantics with deep knowledge.} 
\label{fig:domain}
\end{figure}
\begin{figure*}[t!]
\centering
\includegraphics[width=\linewidth]{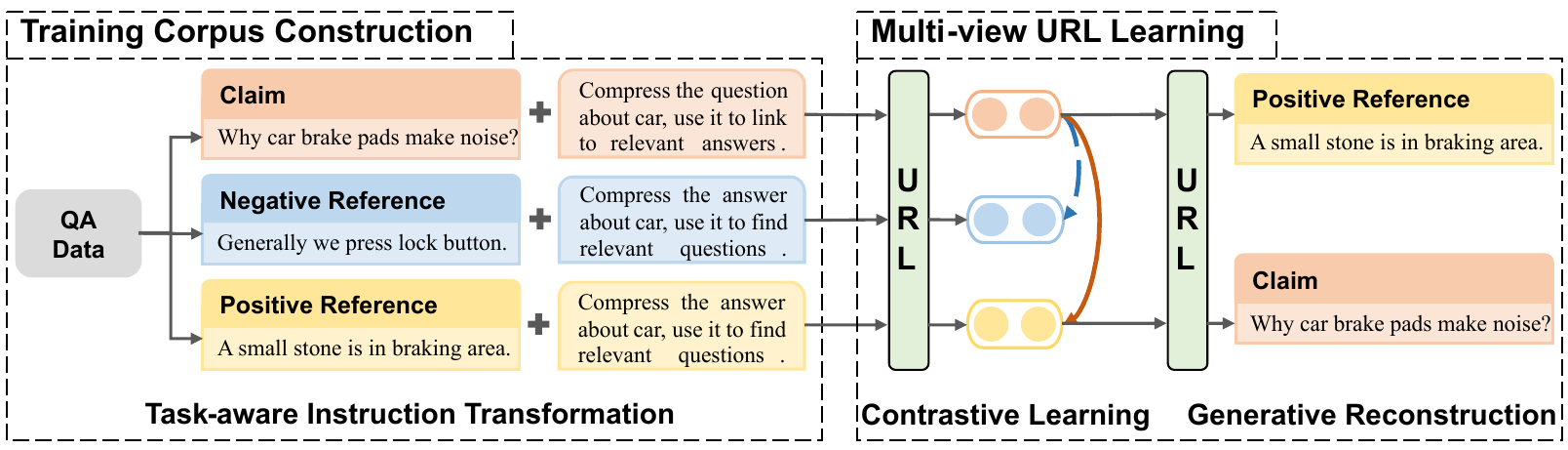} 
\caption{Illustration of training corpus construction and multi-view URL learning. Based on QA data, we set the question as claim and answer as reference, then annotate instructions that describe the field of data and the purpose of representation. For learning, contrastive learning is on embeddings of claims and references, generative reconstruction is to force the model generating positive reference based on claim embedding and vice versa.}
\label{fig:method}
\end{figure*}
Currently, the work on referential knowledge linking is often confined to specific tasks and contexts. As illustrated in Figure~\ref{fig:domain}, information retrieval ~\cite{10184013, muennighoff2023mteb, xiao2023cpack} focuses on locating documents that contain answers to a query, primarily concentrating on the ``contains-answer'' type of relationship. Semantic matching, on the other hand, aims to identify pairs of text segments that share the same semantics~\cite{cer-etal-2017-semeval, 10.1145/3440755}, essentially looking for segments that are ``semantically equivalent''. However, the scenarios for referential knowledge linking in reality are far more diverse. Depending on the particular claim and reference, the objectives of linking can vary significantly. For instance, given a legal case, one might aim to link it to corresponding legal provisions to find the basis of the verdict. Conversely, given a patient symptom, it could be linked to relevant drugs, treatment plans, or possible cause, depending on the specific need.

Compared to the previous focus on specific contexts such as information retrieval or semantic matching, it is challenging to universally solve referential knowledge linking in diverse real-world scenarios. Firstly, although all under the paradigm of RKL, the target semantics of RKL vary across different contexts, i.e., whether a claim and a reference constitute a linking pair may differ under different RKL contexts. For example, given a legal case and a provision, the decision on whether they should be linked may vary drastically under different relational contexts like ``applying for adults'' or ``applying for minors''. Secondly, unlike information retrieval and semantic matching, which often only require superficial semantic understanding (e.g., word matching features and shallow semantic representations), many RKL tasks necessitate deep semantic comprehension and reasoning. Therefore, RKL demands models with robust semantic understanding and, often potentially, reasoning capabilities. Finally, given the typically large size of reference databases, it is impractical to directly perform complex, deep semantic interactions and matching across all claim-reference pairs. Thus, constructing and implementing an efficient, highly adaptable, and semantically capable general RKL model poses a significant challenge.

In this paper, we introduce \textit{\textbf{unified referential linking (URL)}}, a universal framework for linking claims to references on versatile semantics. Specifically, URL leverages the semantic comprehension and instruction-following capabilities of large language models (LLMs) to facilitate universal referential knowledge linking across diverse contexts. However, applying LLMs to RKL poses several challenges due to differences in training objectives and usage modalities. Firstly, the high computational cost of large language models makes direct pair-wise comparison for claim-reference decision-making highly inefficient. Secondly, since large language models are trained in a language model paradigm, directly employing them for universal referential knowledge linking may lead to mismatches in patterns, thereby significantly impacting the performance of LLMs in general RKL tasks. Addressing these key challenges is crucial for harnessing the capabilities of large language models for referential knowledge linking.

To this end, URL introduces a LLM-driven task-instructed representation compression mechanism, which adaptively integrates task-specific information with claim/reference and converts them into vector representations. This approach allows the representations of claims and references to be efficiently adapted across different task scenarios. Levering these vectorized representations, the linking between claims and references can be conducted by directly calculating the association between their task-aware representations, thereby significantly reducing the need for complex computations with large models and enhancing the efficiency of RKL. To facilitate this, we propose a parameter-efficient, multi-view learning algorithm that enables joint training of the large models in both generation and linking modes under the constraint of only observing compressed representations. As illustrated in Figure~\ref{fig:method}, this approach can effectively learn the compressed vector representation by shifting the mode of large language models from context-aware generation to compressed representation-aware generation and linking. Moreover, to train a generalizable and universal RKL representation model effectively, we start with existing question answering (QA) corpus. By transforming QA data from various domains into diverse claim-reference datasets under different linking relationships, we align the model from a language generation mode to a representation compression mode, thereby learning better task-aware RKL representation models.

To validate the effectiveness of universal referential knowledge linking, we introduce a new benchmark—\textit{\textbf{URLBench}}. The primary objective of URLBench is to construct an evaluation set that covers multiple scenarios of referential knowledge linking, thereby assessing the ability of model in generalized referential knowledge linking. Specifically, URLBench encompasses four distinct domains: finance, law, medicine, and education. Within these domains, URLBench evaluates the model capability to link claims and references under various semantic conditions, which primarily includes linking government policy to company profile, legal case to legal provision, patient symptom to drug description, and training objective to course introduction. Experimental results on URLBench demonstrate that URL is an effective approach for achieving universal RKL. Its performance not only surpasses previous models trained on large-scale retrieval and semantic matching datasets, but also significantly outperforms proprietary large-model-based embedding models like OpenAI Text Embedding. This validates the efficacy of the approach proposed in this paper for RKL tasks\footnote{Our code and data are openly available at \url{https://github.com/Li-Z-Q/URL}}.

The main contributions of this paper are:
\begin{itemize}
\setlength{\itemsep}{0.8pt}

\item We define the task of universal referential knowledge linking, which extends beyond considering only a few specific relationships to achieve generalized RKL abilities across diverse scenarios.

\item We propose task-instructed representation compression, a novel framework adapting LLMs to RKL tasks. Meanwhile, we propose a new method based on multi-view RKL learning, which can effectively finetune LLMs by existing QA data.

\item We construct URLBench, a new benchmark across versatile knowledge-rich tasks in various fields, which can effectively evaluate the ability of models in universal referential knowledge linking.
\end{itemize}
\section{Related Work} 

\subsection{Semantic Matching}
Semantic matching is to measure the semantic similarity between two blocks of text~\cite{10.1145/3440755}. 
This is a traditional and fundamental task in natural language processing, containing datasets across many domains and language~\cite{agirre-etal-2012-semeval, agirre-etal-2013-sem, agirre-etal-2014-semeval, agirre-etal-2015-semeval, agirre-etal-2016-semeval, marelli-etal-2014-sick, cer-etal-2017-semeval}. 
However, the relationship in this task is single and static, which is merely about shallow semantic of two sentences, judging whether that ``have similar meaning''.

\subsection{Information Retrieval}
Information retrieval (IR) is the process of searching and returning relevant documents for a query from a collection~\cite{10184013, muennighoff2023mteb}. 
IR contains datasets in various fields and task formats such as question-answer, title-passage, query-document~\cite{thakur2021beir, kwiatkowski-etal-2019-natural, cohan-etal-2020-specter, xiao2023cpack, thorne-etal-2018-fever}. 
However, the essence of all the IR tasks is identical, which is judging if the claim and reference ``contain similar information''. In other words, the relationship in IR is still lacked of universality.

\subsection{Sentence Embedding}
Early embedding methods are based on BERT-style models~\cite{reimers-gurevych-2019-sentence}. Then some works use contrastive learning and get improvement~\cite{gao-etal-2021-simcse,izacard2022unsupervised}. Recent common methods are using two-stage contrastive learning by large scale corpus~\cite{Wang2022TextEB,xiao2023cpack,li2023general}, based on special trained RetroMAE~\cite{xiao-etal-2022-retromae}. And some works also use task instructions on BERT-style models~\cite{su-etal-2023-one}. 

Recently, some works use LLMs to generate sentence embeddings by instruction~\cite{jiang2023scaling}, and contrastive learning~\cite{ma2023finetuning, zhang2023language}. And embeddings generated by LLMs can be used to do retrieval~\cite{wang2023improving, li2023making}, or compression~\cite{gupta2023gistscore, mu2023learning, chevalier-etal-2023-adapting}. 

The important distinction between this work and previous LLM embeddings is that URL focus on addressing versatile RKL tasks from a unified perspective, which receives less attention before. Additionally, beyond contrastive learning, this work proposes a multi-view learning approach.
\section{Universal RKL via Task-instructed Representation Compression} 

Universally addressing RKL tasks poses significant challenges because of versatile linking relationships and deep semantic of claim-reference pairs. Firstly, the relationship between claims and references is flexible across different contexts, requiring varied semantic abilities for linking. Secondly, the relationship involves deep semantic correlations and even some reasoning, demanding deep knowledge and reasoning capabilities from the model. Moreover, handling large-scale reference datasets directly by complex semantic processing on each claim-reference pair causes large time costs. 

In this paper, we propose a novel framework for universal RKL tasks based on LLMs with powerful knowledge, semantic understanding and universality. But applying LLMs presents several challenges due to computational cost and original training objective. Firstly, direct pairwise comparison by LLMs is inefficient, exacerbating efficiency issues with large-scale reference databases. Secondly, since LLMs are trained as language models, directly applying to RKL encounters pattern mismatching. To tackle mentioned challenges, URL employs a task-instructed representation compression mechanism driven by LLMs. The method dynamically incorporates task-specific details with claim/reference data, generating vector representations. To facilitate LLMs for this, this paper introduces a multi-view learning algorithm to enable simultaneous training in both generation and linking modes, on only compressed representations.

Specifically, by LLM-driven task-instructed representation compression, URL combines task information with claims/references and gets task-instructed embeddings, achieving efficient adaptation across different scenarios. In finetuning, multi-view learning injects knowledge of LLMs into  representations and aligns representations with claim-reference embedding-similarity paradigms. To construct training data, URL leverages existing question answering data, transforming QA data from different domains into claim-reference data. This data construction method is a convenient way to simulate real versatile RKL tasks.

\subsection{Task-instructed Compression for RKL}
\label{method:instruction} 

Applying LLMs to practical RKL tasks faces two major challenges: adaptability and efficiency. In adaptability, linking from claim to reference requires diverse semantic abilities depending on the context or application, necessitating a model capable of adapting to various contexts or tasks. In efficiency, there is large computation cost because of the low operating speed of LLMs and possible large scale of reference databases, requiring some efficient linking better than direct comparison. In this paper, URL addresses mentioned challenges by adding task-aware instructions to claims and references, enabling the model to generate task-instructed representations. By computing vectorized representations similarity, URL achieves efficient and adaptive linking for versatile RKL tasks.

Specifically, task-instructed compression compresses the claim/reference, as well as the instruction of target task to be performed into the same vectorized representation, thereby constructing a task oriented vectorized representation: 
\begin{align*}
\mathbf{H}&=\text{LLM}(\boldsymbol{x}_{s},\boldsymbol{x}_{ins},\boldsymbol{x}_{suffix}) \\
\mathbf{e}_{s}&=\text{Pooling}(\mathbf{H}_{suffix})
\end{align*} where $\boldsymbol{x}_{s}$ is the claim or reference, $\boldsymbol{x}_{ins}$ is the task-aware instruction adapted with the claim or reference, $\boldsymbol{x}_{suffix}$ means some suffix tokens, $\mathbf{H}$ is  last-layer hidden states of all tokens and $\mathbf{H}_{suffix}$ is that of suffix tokens,  $\mathbf{e}_{s}$ is the output embedding. 

In summary, by calculating vector similarity of task-instructed representations of the claim and reference, URL gets the linking score of claim-reference pairs, and address RKL tasks efficiently and universally without complex computation. 

\subsection{Multi-view URL Learning}
\label{sec:multi_view}

Under the aforementioned URL framework, a core issue is how to effectively learn task-instructed representation for versatile RKL tasks. Given that LLMs are primarily designed for language generation tasks, directly applying representations from LLMs to RKL tasks is challenging. And solely employing claim-reference pairs contrastive learning presents damaging the original knowledge and reasoning capabilities obtained during pre-training. In this paper, we propose a novel multi-view learning approach that facilitates simultaneous generative reconstruction and contrastive learning processes, enabling large models to swiftly and cost-effectively adapt to URL norms while preserving existing knowledge and reasoning abilities. 

Specifically, multi-view URL learning comprises two components: generative reconstruction and contrastive learning. The generative reconstruction employs claim embeddings to generate relevant references and reference embeddings to generate relevant claims, enabling the injection of model knowledge  into compressed representations to better address reasoning-style linking tasks in RKL. The second component, contrastive learning, minimizes the distance between related claim and reference representations while maximizing the distance between unrelated ones, enabling model-generated representations to align with the embedding-similarity paradigm.
\paragraph{Generative Reconstruction.} By generative loss based on the claim representation with relevant reference, and reference representation with relevant claim, generative reconstruction injects knowledge of the LLM into the compressed representation: 
\begin{equation}
\mathcal{L}_1=\sum_{\mathcal{D}}^{}-\log p(\boldsymbol{x}_{pos}|\mathbf{e}_c,\boldsymbol{x}_{p})
\nonumber 
\end{equation} where $\mathcal{D}$ means training data, $\mathbf{e}_c$ is the claim embedding, $\boldsymbol{x}_{pos}$ is the positive reference, $\boldsymbol{x}_{p}$ is the prompt used to guide the LLM to generate the relevant reference. On another hand, reconstructing claim based on reference embedding has symmetric formula as above. Through generative reconstruction, the representation contains great knowledge and deep understanding of the input sentence because the LLM can generate relevant contents merely based on the compressed representation rather than original complete sentences. 
\paragraph{Contrastive Learning.} By minimizing the distance between representations of the relevant claim and reference and maximizing that of the irrelevant claim and reference, contrastive learning directly train the LLM for suiting the claim-reference embedding-similarity pattern for URL:
\begin{equation}
\mathcal{L}_2=\sum_{\mathcal{D}}^{}-\log \frac{e^{\left \langle \mathbf{e}_{c},\mathbf{e}_{pos} \right \rangle /\tau } }{e^{\left \langle \mathbf{e}_{c},\mathbf{e}_{pos} \right \rangle /\tau } +  \sum_{\mathbf{e}_{neg}}^{} e^{\left \langle \mathbf{e}_{c},\mathbf{e}_{neg} \right \rangle /\tau } }
\nonumber 
\end{equation} where $\mathbf{e}_{pos}$ is embedding of the positive reference, $\mathbf{e}_{neg}$ is for the negative reference, and $\tau$ is the temperature parameter. 
Through contrastive learning, the LLM are trained to efficiently address  RKL tasks by embedding-similarity method. 
\paragraph{Multi-view Learning.} By composing the loss of generative reconstruction and contrastive learning, multi-view URL learning integrates the advantages of two training methods, the LLM can be well aligned for versatile knowledge-rich RKL tasks:
\begin{equation}
\mathcal{L}_{total}=\alpha\mathcal{L}_1 + (1-\alpha)\mathcal{L}_2
\nonumber 
\end{equation} where $\alpha$ is a parameter. In actual training, URL uses symmetric bi-direct loss as some recent works~\cite{xiao2023cpack}. Respectively, for generative reconstruction, URL also calculates loss by generating the claim based on embedding of the correct reference, for contrastive learning, URL also calculates loss among embeddings of one reference, the relevant claim, and irrelevant claims. 

In summary, multi-view URL learning ensures task-instructed representations encompass knowledge of the LLM, and suit for the claim-reference embedding-similarity pattern. The total training process can effectively align the LLM for universally addressing versatile RKL tasks. 

\subsection{Constructing URL Training Data via QA Corpus Transformation}
\label{method:data}

In order to support the multi-view URL learning mentioned above, a core issue is to construct training data for versatile RKL tasks. This presents a crucial challenge as existing works primarily focus on information retrieval and semantic matching, both of which entail singular linking relationships, lacking diverse training data for versatile linking tasks. In this paper, we propose constructing a URL training corpus by transforming QA data. By annotating instructions for ordinary question-answer data according to the domain of data, this construction method utilizes versatile domain data to simulate versatile tasks in RKL.

\begin{table*}[htbp]
\centering
\resizebox{\linewidth}{!}{
    \begin{tabular}{ccccccrcccc}
\toprule
    \multirow{2}[4]{*}{\textbf{Domain}} & \multirow{2}[4]{*}{\textbf{Application}} & \multirow{2}[4]{*}{\textbf{Relationship}} & \multicolumn{4}{c}{\textbf{Claim}} & \multicolumn{4}{c}{\textbf{Reference}} \\
\cmidrule(lr){4-7}  \cmidrule(lr){8-11}    

&       &       & \textbf{Content} & \textbf{Number} & \multicolumn{2}{c}{\textbf{Length}} & \textbf{Content} & \textbf{Number} & \multicolumn{2}{c}{\textbf{Length}} \\
    \midrule
    Finance &    Stock  Decision   & Affecting & Government Policy & 1000  & 157   & 252   & Company Profile & 709   & 1026  & 922 \\
    Law   &    Legal  Judgement   & Confirming & Legal Case & 1000  & 236   & 390   & Legal Provision & 3627  & 56    & 103 \\
    Medicine &  Medical Prescription     & Treating & Patient Symptom & 750   & 80    & 148   & Drug Description & 1000  & 36    & 70 \\
    Education &   Course Planning    & Supporting & Training Objective & 133   & 90    & 145   & Course Introduction & 787   & 115   & 153 \\
    \bottomrule
    \end{tabular}%
}
  \caption{Detailed statistics of the benchmark. Length is the average token number of sentences by LLM tokenizers, the left value is for Chinese version and the right is for English version.}
  \label{tab:statistics}
\end{table*}%
Specifically, for 1000 question-answer pairs selected from mMARCO~\cite{bonifacio2022mmarco}, we manually categorize data into 40 domains, annotate instructions for each domain, and set the question as claim and the answer as reference. As illustrated in Figure~\ref{fig:method}, each data includes a claim, a positive reference, some negative references by random sampling, and two simulating task-aware instructions. In addition, this training dataset has versions for both Chinese and English language. 

In summary, the method uses heterogeneous domains to simulate versatile relationships of RKL. This is an efficient alternative solution in the situation absenting real versatile training data for RKL. 

\section{URLBench: Benchmarking Universal Referential Knowledge Linking} 
\label{sec:benchmark}

Currently, RKL benchmarks primarily focus on two specific tasks: semantic matching and information retrieval. The former emphasizes ``having similar semantics'', while the latter focuses on ``containing similar information''. However, RKL tasks are versatile and with challenging knowledge-rich relationships in real-world applications, there is a lack of a benchmark that covers versatile RKL tasks and reflects complex deep linking relationships, posing a significant obstacle to evaluating unified models. In this paper, we propose a new benchmark that encompasses various RKL tasks across multiple domains and real applications. The benchmark can greatly reflect the diversity and knowledge-rich challenge of RKL tasks.

Specifically, by focusing on  four real-world applications including stock decision, legal judgement, medical prescription and course planning, URLBench collects data from web or existing datasets to create four evaluation tasks: linking government policies and company profiles, legal cases and legal provisions, patient symptoms and drug descriptions, and training objectives and course introductions. 
Focusing on stock decision, URLBench contains linking from government policies to companies that can be affected, constructed by transforming an existing task~\cite{wang2023csprd} to policy-company linking format.
Focusing on legal judgement, URLBench contains linking from legal cases to provisions that can confirm the case, constructed by transforming and extending an existing classification  task~\cite{xiao2021lawformer} referring to web\footnote{http://www.law-lib.com/}.
Focusing on medical prescription, URLBench contains linking from patient symptoms to drugs that can treat the symptom, constructed by transforming and extending an existing classification task~\cite{he-etal-2022-dialmed} referring to web\footnote{https://zyp.yilianmeiti.com}.
Focusing on course planning, URLBench contains linking from training objectives to courses that can support the objective. This task is constructed based on a university enrollment handbook\footnote{https://dean.pku.edu.cn/web/download.php}.

Clear statistics are shown in Table~\ref{tab:statistics}, the scale of the benchmark can ensure experiments quick and effective, and the length of datas is suitable for most models. Note that these datasets are originally mainly in Chinese language, we do translation by gpt-3.5-turbo\footnote{https://platform.openai.com/docs/guides/text-generation\label{foot:3.5}} to get English version. We then do some filter and checking to ensure the quality.
\section{Experiments}
\subsection{Experimental Settings}
\label{sec:settings}
We choose LLAMA-2-7B-Chat~\cite{touvron2023llama} as the base model for English tasks, and  Baichuan-2-7B-Chat~\cite{yang2023baichuan} for Chinese tasks. We finetune LLMs by the 1000 training data by LoRA method~\cite{hu2021lora} in multi-view learning, setting lora\_r as 8, lora\_alpha as 16 and lora\_dropout as 0.05. Respectively, we set lora\_target as ``W\_pack'' for Baichuan model and ``q\_proj v\_proj'' for LLAMA model. Thanks to these settings, we finetune 7B-size LLMs on one A-100 80G GPU. The total finetuning process has 62 training steps and needs about 20 minutes.

We choose NDCG~\cite{pmlr-v30-Wang13} and MAP~\cite{carterette2011overview} as metrics in evaluating. And we utilize the Python interface of the official TREC evaluation tool~\cite{van2018pytrec_eval} to ensure the reliability of  results. 

\begin{table*}[t!]
\centering
\resizebox{\linewidth}{!}{
\begin{tabular}{lcccccccccccccccc}
\toprule

\multirow{3}[1]{*}{\textbf{Method}} & \multicolumn{4}{c}{\textbf{Policy-Company}} & \multicolumn{4}{c}{\textbf{Case-Provision}} & \multicolumn{4}{c}{\textbf{Symptom-Drug}} & \multicolumn{4}{c}{\textbf{Objective-Course}} \\

\cmidrule(lr){2-5} \cmidrule(lr){6-9}  \cmidrule(lr){10-13} \cmidrule(lr){14-17}
& \multicolumn{2}{c}{\textbf{NDCG}} & \multicolumn{2}{c}{\textbf{MAP}} & \multicolumn{2}{c}{\textbf{NDCG}} & \multicolumn{2}{c}{\textbf{MAP}} & \multicolumn{2}{c}{\textbf{NDCG}} & \multicolumn{2}{c}{\textbf{MAP}} & \multicolumn{2}{c}{\textbf{NDCG}} & \multicolumn{2}{c}{\textbf{MAP}} \\
  
\cmidrule(lr){2-3} \cmidrule(lr){4-5}  \cmidrule(lr){6-7} \cmidrule(lr){8-9} \cmidrule(lr){10-11} \cmidrule(lr){12-13}  \cmidrule(lr){14-15} \cmidrule(lr){16-17}
& \textbf{@10} & \textbf{@20} & \textbf{@10} & \textbf{@20} & \textbf{@10} & \textbf{@20} & \textbf{@10} & \textbf{@20} & \textbf{@10} & \textbf{@20} & \textbf{@10} & \textbf{@20} & \textbf{@10} & \textbf{@20} & \textbf{@10} & \textbf{@20} \\
\midrule
    \rowcolor[rgb]{ .906,  .902,  .902} \multicolumn{17}{c}{English Evaluation} \\
    \midrule
    BM25  & 36.3  & 39.3  & 27.3  & 28.7  & 2.9   & 3.8   & 1.5   & 1.9   & 4.6   & 6.4   & 2.7   & 3.2   & 22.0  & 24.9  & 11.1  & 13.1  \\
    Voyage-lite-02-instruct & 43.9  & 47.5  & 34.0  & 35.9  & 12.8  & 14.7  & 6.7   & 7.8   & 10.7  & 12.8  & 7.0   & 7.6   & 38.0  & 41.0  & 22.6  & 25.6  \\
    Text-embedding-ada-002 & 46.7  & 50.3  & 36.7  & 38.7  & 9.3   & 11.9  & 6.5   & 7.9   & 11.1  & 13.5  & \underline{7.3}   & \underline{8.1}   & 38.7  & 42.5  & 23.7  & 27.4  \\
    Text-embedding-3-large & \textbf{49.3} & \textbf{53.0} & \textbf{39.3} & \textbf{41.3} & \underline{20.1}  & \underline{21.0}  & \underline{14.2}  & \underline{14.4}  & 10.2  & 12.7  & 6.1   & 6.8   & \underline{44.3}  & \underline{48.0}  & \underline{28.0}  & \underline{32.2}  \\
    E5-7B~\cite{wang2023improving} & 41.8  & 45.7  & 32.5  & 34.6  & 18.5  & 20.2  & 12.7  & 13.7  & \underline{11.9}  & \underline{15.1}  & 7.2   & \underline{8.1}   & 39.1  & 42.7  & 23.2  & 26.7  \\
    UAE~\cite{li2023angleoptimized}   & 46.6  & 50.2  & 36.4  & 38.3  & 5.9   & 7.5   & 3.1   & 4.0   & 8.4   & 11.0  & 5.2   & 6.0   & 42.1  & 46.2  & 26.8  & 30.6  \\
    BGE~\cite{xiao2023cpack}   & 43.6  & 47.7  & 33.8  & 35.8  & 9.5   & 11.2  & 5.8   & 6.6   & 8.2   & 10.0  & 5.1   & 5.6   & 39.3  & 43.2  & 24.2  & 27.7  \\
    GTE~\cite{li2023general}   & 46.4  & 50.0  & 36.4  & 38.2  & 14.7  & 16.6  & 10.2  & 11.4  & 9.4   & 12.2  & 6.0   & 6.8   & 42.1  & 45.9  & 26.2  & 30.0  \\
    E5~\cite{Wang2022TextEB}    & 38.8  & 42.6  & 29.6  & 31.4  & 7.4   & 8.5   & 5.2   & 5.9   & 6.4   & 8.7   & 4.0   & 4.5   & 34.7  & 38.9  & 21.0  & 24.3  \\
    INSTRUCTOR~\cite{su-etal-2023-one} & 45.5  & 49.6  & 35.6  & 37.8  & 17.5   & 19.1   & 12.2   & 12.6   & 10.0   & 13.0   & 6.1   & 7.0   & 37.3  & 42.2  & 22.2  & 26.2 \\
    \textbf{URL} (ours) & \underline{47.1}  & \underline{51.0}  & \underline{37.5}  & \underline{39.7}  & \textbf{29.3} & \textbf{30.4} & \textbf{20.2} & \textbf{21.3} & \textbf{13.8} & \textbf{16.1} & \textbf{9.7} & \textbf{10.4} & \textbf{48.2} & \textbf{53.4} & \textbf{31.1} & \textbf{37.4} \\    \midrule
    \rowcolor[rgb]{ .906,  .902,  .902} \multicolumn{17}{c}{Chinese Evaluation} \\
    \midrule
    BM25  & 23.7  & 26.3  & 17.4  & 18.4  & 10.2  & 11.1  & 8.0   & 8.3   & 5.5   & 7.1   & 3.4   & 3.9   & 18.7  & 20.9  & 10.4  & 11.7  \\
    Baichuan-text-embedding & 26.6  & 30.5  & 18.8  & 20.4  & \underline{25.0}  & \underline{26.9}  & \underline{16.9}  & \underline{17.5}  & 10.3  & 12.6  & 6.7   & 7.3   & 16.0  & 17.5  & 7.2   & 8.2  \\
    Text-embedding-ada-002 & \underline{47.3}  & \underline{51.3}  & \underline{37.2}  & \underline{39.4}  & 17.5  & 19.9  & 10.5  & 11.2  & 7.0   & 9.1   & 4.3   & 4.8   & 34.7  & 38.4  & 20.7  & 23.8  \\
    Text-embedding-3-large & 45.8  & 49.6  & 35.7  & 37.7  & 24.2  & 26.6  & 16.1  & 16.9  & 10.2  & 13.3  & 6.0   & 6.9   & \underline{42.8}  & \underline{46.4}  & \underline{27.1}  & \underline{31.1}  \\
    E5-7B~\cite{wang2023improving} & 37.1  & 41.2  & 27.3  & 29.3  & 16.7  & 18.6  & 10.3  & 10.7  & 10.6  & 13.8  & 6.8   & 7.7   & 40.3  & 43.6  & 25.0  & 28.3  \\
    GTE~\cite{li2023general}   & 32.9  & 36.7  & 24.5  & 26.2  & 7.7   & 9.1   & 4.3   & 4.6   & 10.2  & 13.6  & 6.3   & 7.2   & 40.3  & 43.6  & 25.2  & 28.5  \\
    BGE~\cite{xiao2023cpack}   & 38.7  & 42.8  & 29.2  & 31.2  & 24.3  & 26.5  & 16.4  & 17.1  & \underline{11.6}  & \underline{14.1}  & \underline{7.1}   & \underline{7.8}   & 38.3  & 41.2  & 22.5  & 25.6  \\
    \textbf{URL} (ours) & \textbf{49.4} & \textbf{53.0} & \textbf{39.3} & \textbf{41.5} & \textbf{31.0} & \textbf{33.1} & \textbf{21.1} & \textbf{21.9} & \textbf{14.7} & \textbf{17.2} & \textbf{9.8} & \textbf{10.5} & \textbf{50.5} & \textbf{55.1} & \textbf{32.9} & \textbf{38.9} \\
    \bottomrule
    \end{tabular}%
}
\caption{Main experiments in URLBench. Best/second-best performing score in each column is highlighted with bold/underline. It shows that URL is with better performance and universality than baselines.}
\label{tab:main_results}%
\end{table*}%

\begin{table}[t]
\centering
\resizebox{\linewidth}{!}{
    \begin{tabular}{lcccc}
    \toprule
    \textbf{Method} & \textbf{Policy-C.} & \textbf{Case-P.} & \textbf{Symptom-D.} & \textbf{Objective-C.} \\
    \midrule
    \textbf{URL} (ours) & \textbf{47.1} & \textbf{29.3} & \textbf{13.8} & \textbf{48.2} \\
    \midrule
    \rowcolor[rgb]{ .906,  .902,  .902} \multicolumn{5}{l}{Further Finetuning BERT-style Models} \\
    \midrule
    BGE~\cite{xiao2023cpack}   & 36.7$\downarrow$  & ~~7.8$\downarrow$   & ~~7.8$\downarrow$   & 40.9$\uparrow$ \\
    GTE~\cite{li2023general}    & 35.6$\downarrow$  & 12.1$\downarrow$  & ~~7.1$\downarrow$   & 42.6$\uparrow$ \\
    \midrule
    \rowcolor[rgb]{ .906,  .902,  .902} \multicolumn{5}{l}{Using Task-aware Instructions on API} \\
    \midrule
    Text-embedding-ada-002   & 36.9$\downarrow$  & ~~7.8$\downarrow$   & ~~10.1$\downarrow$   & 35.6$\downarrow$ \\
    Text-embedding-3-large   & 35.0$\downarrow$  & 19.6$\downarrow$  & ~~11.4$\uparrow$   & 42.4$\downarrow$ \\
    \bottomrule
    \end{tabular}%
}
\caption{NDCG@10 scores for further finetuning BERT-style models by our training data and using task-aware instructions on powerful API. It shows some fluctuations, yet still significantly under-performs URL.}
\label{tab:train_encoder}%
\end{table}%
\subsection{Baselines}
Referring to MTEB leader board\footnote{https://huggingface.co/spaces/mteb/leaderboard\label{foot:mteb}}, we select some top-ranking and common used models as baseline, including a traditional sparse retriever BM25~\cite{robertson2009probabilistic}, four powerful closed source API, text-embedding-ada-002 and text-embedding-3-large\footnote{https://platform.openai.com/docs/guides/embeddings\label{foot:002}}, baichuan-text-embedding\footnote{https://platform.baichuan-ai.com/docs/text-Embedding}, and voyage-lite-02-instruct\footnote{https://docs.voyageai.com/embeddings/}, some open source embedding models trained by large scale retrieval datas, uae-large-v1~\cite{li2023angleoptimized}, gte-large and gte-large-zh~\cite{li2023general}, bge-large-en-v1.5 and  bge-large-zh-v1.5~\cite{xiao2023cpack}, e5-large-v2~\cite{Wang2022TextEB}, instructor-xl~\cite{su-etal-2023-one}. Additionally, we also evaluate e5-mistral-7b-instruct~\cite{wang2023improving}, which is a LLM embedder trained by contrastive learning and expensive high-quality multi-language datas.

\begin{table*}[t!]
  \centering
\resizebox{\linewidth}{!}{
    \begin{tabular}{lcccccccccccc}
    \toprule
    \multirow{2}[1]{*}{\textbf{Method}} & \multicolumn{4}{c}{\textbf{Components}} & \multicolumn{2}{c}{\textbf{Policy-Company}} & \multicolumn{2}{c}{\textbf{Case-Provision}} & \multicolumn{2}{c}{\textbf{Symptom-Drug}} & \multicolumn{2}{c}{\textbf{Objective-Course}} \\
    
   \cmidrule(lr){2-5} \cmidrule(lr){6-7} \cmidrule(lr){8-9} \cmidrule(lr){10-11} \cmidrule(lr){12-13} 
    
    & \textbf{Task-aware} & \textbf{Instruction} & \multicolumn{1}{p{5.78em}}{\textbf{Generative}} & \multicolumn{1}{p{6.11em}}{\textbf{Contrastive}} & \textbf{NDCG} & \textbf{MAP} & \textbf{NDCG} & \textbf{MAP} & \textbf{NDCG} & \textbf{MAP} & \textbf{NDCG} & \textbf{MAP} \\
          
    \midrule
          
    \textbf{URL} (ours) &  \cmarkgreen     &  \cmarkgreen     &  \cmarkgreen     &   \cmarkgreen    & 47.1  & 37.5  & 29.3  & \textbf{20.2} & \textbf{13.8} & \textbf{9.7} & \textbf{48.2} & \textbf{31.1} \\
    \midrule
    \rowcolor[rgb]{ .906,  .902,  .902} \multicolumn{13}{l}{Ablating Task-aware Instruction} \\
    \midrule
    w/o Instruction &  \xmarkred     &  \xmarkred     &  \cmarkgreen     &   \cmarkgreen  & 37.7  & 28.4  & 23.1  & 15.3  & 5.7   & 3.1   & 24.6  & 13.0  \\
    w/o Task-aware Instruction &  \xmarkred     &  \cmarkgreen     &  \cmarkgreen     &   \cmarkgreen    & \textbf{48.4} & \textbf{38.8} & \textbf{29.7} & 19.7  & 11.7  & 8.1   & 40.0  & 24.5  \\
    \midrule
    \rowcolor[rgb]{ .906,  .902,  .902} \multicolumn{13}{l}{Ablating Multi-view Learning} \\
    \midrule
    w/o Learning &  \cmarkgreen     &  \cmarkgreen     &  \xmarkred     &   \xmarkred   & 19.3  & 13.1  & 4.5   & 2.9   & 5.3   & 3.6   & 28.1  & 16.4  \\
    w/o Generative Reconstruction &  \cmarkgreen     &  \cmarkgreen     &  \xmarkred     &   \cmarkgreen    & 46.1  & 36.6  & 28.2  & 19.7  & 12.7  & 8.9   & 46.4  & 29.1  \\
    \bottomrule
    \end{tabular}%
}
    \caption{NDCG@10 and MAP@10 results of ablations. It shows that task-aware instructions and multi-view learning are great helpful for addressing RKL tasks universally and effectively.}
    \label{tab:ablations}%
\end{table*}%

\subsection{Overall Results}
In order to validate the performance of URL on versatile RKL tasks and illustrate the necessity of URL, experiments are conducted based on URLBench from three perspectives, including comparing performance of URL and baselines, further finetuning BERT-style models by our 1000 training data, and using task-aware instructions on API. Experiments show that URL is universal and accurate, simple finetuning on BERT-style models and direct adding instructions on API are not good solutions.

Firstly, by comparing the performance of open source embedding models, powerful closed-source API, and URL on the benchmark, as shown in Table~\ref{tab:main_results}, results demonstrate that although some baseline models achieve similar performance as URL on the certain dataset, none of them can consistently perform well across all the versatile tasks. The conclusion is that URL can perform high performance and great universality for RKL tasks. 

Secondly, by observing performance of original BERT-style models trained further on our training data, as shown in Table~\ref{tab:train_encoder}, it shows some performance fluctuations, but BERT-style models are surely not enough to address universal RKL as URL. Conclusion is that BERT-style models cannot achieve same level of universal high performance as URL with only 1000 training data samples. 

Finally, by observing the performance of API with task-aware instructions, as shown in Table~\ref{tab:train_encoder}, scores of both  text-embedding-ada-002 and text-embedding-3-large become worse in most tasks when directly using task-aware instructions on it. The conclusion is that directly using API cannot achieve good and universal performance. 

In summary, through the aforementioned three perspectives of experimentation, the results show that URL is a reliable unified solution, and the framework is valuable and necessary for the development of RKL area because existing BERT-style models and powerful closed API cannot achieve performance close to URL through simple further training or task instructions.

\subsection{Ablations for Task-aware Instruction and Multi-view Learning}

Note that task-aware instruction and multi-view learning are two important modules. To validate their impact for URL framework,  we conduct two types of ablation experiments about instructions and learning, results are shown in Table~\ref{tab:ablations}.

Firstly, by observing performance when no instructions are used and when task-aware instructions are not employed, with only a fixed instruction, it shows that using a single instruction is better than using none at all, but  clearly not as universal as task-aware instructions. Conclusion is instructions significantly affect performance, and task-aware instruction is with important effect in URL. 

Secondly, by comparing performance of LLM without training and LLM trained only via contrastive learning, it shows that raw LLMs perform badly, and multi-view learning shows better performance beyond contrastive learning. The conclusion is that original LLMs need training to apply to RKL tasks, and multi-view learning is a more effective training approach than contrastive learning.

In summary, task-aware instructions and multi-view learning show significant effect in URL framework, these two components are helpful to align LLMs with better ability for versatile RKL tasks.

\begin{table}[t]
\centering
\resizebox{\linewidth}{!}{
    \begin{tabular}{p{0.1\textwidth}p{0.32\textwidth}<{\centering}p{0.32\textwidth}<{\centering}}
    \toprule
     & \textbf{Claim} & \textbf{Knowledgeable Reference} \\
    \midrule
    \textbf{Policy-Company} & Establish a number of \textbf{Internet of Things} technology laboratories ... & Company mainly engaged in the research of \textbf{MEMS sensors} ... \\
    \midrule
    \textbf{Objective-Course} & This major focuses on  systematic mastery of \textbf{robot engineering} ... & This course is to introduce the \textbf{feedback control} systems ... \\
    \bottomrule
    \end{tabular}%
}
\caption{BERT-style models fail on these cases because linking is versatile and knowledgeable, rather than with fixed and simple semantic similarity. URL can handle correctly due to powerful knowledge and adaptability.}
\label{tab:case}%
\end{table}%
\subsection{Comparing URL and Conventional Models on Some Cases}
\label{sec:case_study}

To further investigate why URL performs better than conventional models, we analyze some claim-reference pairs that URL deals right but BERT-style models do not. And we find that URL is better mainly because of greater ability for knowledge-rich and versatile cases. 

As shown in Table~\ref{tab:case}, the course about ``feedback control'' can support objective about ``robot engineering'', the company about ``MEMS sensors'' is affected by policy about ``Internet of Things''. These versatile cases are with deep knowledge rather than merely fixed superficial semantic correlation, thus BERT-style models are easy to fail. But URL can handle correctly due to powerful adaptability and knowledgeable semantic understanding ability, further improving multi-view learning is an effective method to align LLMs for RKL.

\section{Conclusion}

In this paper, we first define versatile referential knowledge linking tasks from a unified perspective. To address versatile RKL tasks efficiently and universally, we propose task-instructed representation compression framework and multi-view learning by transformed existing QA data. Furthermore, we introduce a new benchmark across versatile knowledge-rich tasks in various fields. According to the evaluation, URL can address versatile RKL tasks effectively and universally, perform better than powerful OpenAI Text Embedding and common-used open source models. This paper can promote intelligence to generate more authoritative and precise content in the era of LLMs.
\section{Limitations}
In this paper, we primarily investigate the scenario of fine-tuning LLMs with a small amount of data under parameter efficient conditions. We find that the aforementioned small-scale fine-tuning and limited data are sufficient to effectively transfer LLMs for universally addressing referential knowledge linking tasks. What remains unexplored in this paper is whether introducing larger-scale data and finetuning more parameters could achieve an even better model for universal referential knowledge linking tasks. This remains as our future work.

\section{Ethical Considerations}
Our study is with slim chance for ethical risks. On the one hand, we use LLMs for embedding generators rather than content generation. On the other hand, our training and evaluating data are mainly transformed from public safe source. The only possible risk is that we use help of gpt-3.5-turbo  when constructing the benchmark. For this, we conduct manual review to ensure that the content generated by the API does not pose any moral hazard.

\bibliography{custom}

\newpage
\appendix

\section{Detailed Parameters in Multi-view URL Learning}
Here are some supplementary explanation of the finetuning process for Section~\ref{sec:settings}. We finetune LLMs by the 1000 datas for 2 epochs, setting max\_length for both question and answer as 256, batch\_size as 2, gradient\_accumulation\_steps as 16, learning\_rate as 1e-4 and warmup\_ratio as 0.08. For formulas in Section~\ref{sec:multi_view}, we set $\alpha$ as 0.2, $\tau$ as 0.8, and prompt $\boldsymbol{x}_{p}$ as follows:\newline
\textit{Generating reference base on claim embedding:} According to the above text and instructions, these common questions in the Car domain can retrieve the following explanations: \newline
\textit{Generating claim base on reference embedding:} According to the above text and instructions, these explanations in the Car domain can be matched with the following common questions: 

\section{Detailed Processing in Constructing URLBench}
Here are some detailed processing for Section~\ref{sec:benchmark}. 

For policy-company task, existing dataset is a retrieval task about finding related policies based on a given company profile. Original setting can not directly reflect applications in stock decision. In this paper, we directly do some transformation, getting task that linking to related companies based on a given government policy. 

For case-provision task, existing dataset is a classification task, legal case is as text and legal provision is as label. The number of label in original dataset is limited, we manually extend more provisions based on legal web.

For symptom-drug task, existing dataset is a classification task, patient-doctor dialogue is as text and drug name is as label. Firstly, we transform the dialogue to symptom description by the patient because doctor can already talk about the drug. Secondly, we add drug description based on medical web for each drug because drug name contains too little information, is not enough to support linking. Finally, we add more drugs according to the medical web.

For objective-course task, we collect dataset from a university enrollment handbook. Firstly, original file is in PDF format, we transform it to JSON by python tools. Secondly, there are some courses with similar introduction, i.e. Matrix Theory A and Matrix Theory B. We manually recognize these course and only keep one.

\section{Task-aware instructions in Evaluating}
Task-aware instructions for the four evaluation tasks are shown as follows:\newline
\textit{Task-aware instructions for claim:} Above text is description of a \{government policy / legal case / patient symptom / training objective\}. Based on your own knowledge, compress it into an embedding that can be used to search for relevant \{company profile / legal provision / drug description / course introduction\}. The embedding is: \newline
\textit{Task-aware instructions for reference:} Above text is a \{company profile / legal provision / drug description / course introduction\}. Based on your own knowledge, compress it into an embedding that can be used to match relevant \{government policy / legal case / patient symptom / training objective\}. The embedding is:

\begin{table}[t]
\centering
\tiny
\begin{tabular*}{\linewidth}{p{4.5cm}cc}
\toprule[0.5pt]
\textbf{Case} & \textbf{Challenging} & \textbf{Easy} \\
\midrule[0.05pt]
URL Success and BGE~\cite{xiao2023cpack} Fail & 75\% & 25\% \\
URL Success and Only Contrastive Learning Fail & 82\% & 18\% \\
\bottomrule[0.5pt]
\end{tabular*}%
\caption{Two categories of cases that URL can handle correctly but other methods can not, which are claim-reference pairs requiring challenging knowledge or easy knowledge. In these cases, the proportion of challenging data is far greater than shallow datas.}
\label{tab:error}%
\end{table}%
\section{Statistics for Cases URL Can Handle but Others Cannot}
\label{app:case_study}
In order to further investigate why URL outperforms BERT-style models and the role of generative reconstruction, as shown in Table~\ref{tab:error}, we analyze some claim-reference pairs that our model deal right but others do not. With help of gpt-3.5-turbo\textsuperscript{\ref{foot:3.5}}, we annotate each claim-reference pair as data requiring challenging knowledge or easy knowledge, it shows that URL is better mainly because of stronger capabilities in datas requiring challenging knowledge. 

In summary, URL effectively harnesses knowledge and reasoning capabilities of LLMs, and the generative reconstruction greatly keep the embedding with more knowledge when aligning the LLM to RKL tasks, thereby to a certain extent reducing the "alignment tax"~\cite{noukhovitch2023language}. 
 
 \end{document}